\def\year{2021}\relax
\documentclass[letterpaper]{article} 
\usepackage{aaai21}  
\usepackage{times}  
\usepackage{helvet} 
\usepackage{courier}  
\usepackage[hyphens]{url}  
\usepackage{graphicx} 
\usepackage{natbib}  
\usepackage{caption} 
\usepackage{amssymb}
\usepackage[switch]{lineno}
\usepackage{amsmath}

\usepackage{subfigure}
\usepackage{pifont}
\usepackage{multirow}

\newcommand{\cmark}{\ding{51}}%
\newcommand{\xmark}{\ding{55}}%
\newlength\savewidth\newcommand\shline{\noalign{\global\savewidth\arrayrulewidth
  \global\arrayrulewidth 1pt}\hline\noalign{\global\arrayrulewidth\savewidth}}

\title{Unchain the Search Space \\ with Hierarchical Differentiable Architecture Search}
\author {
        Guanting Liu, 
        Yujie Zhong,  
        Sheng Guo,  
        Matthew R. Scott, 
        Weilin Huang\thanks{Corresponding author} \\ 
}
\affiliations {
    Malong LLC \\
    \{gualiu, jaszhong, sheng, mscott, whuang\}@malongtech.com
}
\begin{document}

\maketitle

\begin{abstract}
  Differentiable architecture search (DAS) has made great progress in searching for  high-performance architectures with reduced computational cost.
  However, DAS-based methods mainly focus on searching for a repeatable cell structure, which is then stacked sequentially in multiple stages to form the networks. This configuration significantly reduces the search space, and ignores the importance of connections between the cells.
  To overcome this limitation, in this paper, we propose a Hierarchical Differentiable Architecture Search (H-DAS) that performs architecture search both at the cell level and at the stage level. 
  Specifically, the cell-level search space is relaxed so that the networks can learn stage-specific cell structures. For the stage-level search, we systematically study the architectures of stages, including the number of cells in each stage and the connections between the cells.  
  Based on insightful observations, we design several search rules and losses, and mange to search for  better stage-level architectures.
  Such hierarchical search space greatly improves the performance of the networks without introducing expensive search cost.
  Extensive experiments on CIFAR10 and ImageNet demonstrate the effectiveness of the proposed H-DAS. Moreover, the searched stage-level architectures can be combined with the cell structures searched by existing DAS methods to further boost the performance.
   Code is available at: https://github.com/MalongTech/research-HDAS
\end{abstract}

\section{Introduction}

A large number of neural networks or architectures has been designed for various computer vision tasks in the past years ~\cite{krizhevsky2012imagenet,simonyan2014very,szegedy2015going,he2016deep,huang2017densely}, where human experts played an important role. Such  manually-designed architectures have been proved to be effective, but heavily depend on human skills and experience. 
Recently, Neural Architecture Search (NAS) has attracted increasing attentions~\cite{baker2016designing,zoph2016neural,pham2018efficient}, and achieved the state-of-the-art performance in various computer vision tasks, including image classification~\cite{real2019regularized,tan2019efficientnet} and object detection~\cite{wang2020fcos,zhong2020representation}.

\begin{figure}[t]
  \centering
  \includegraphics[width=0.47\textwidth]{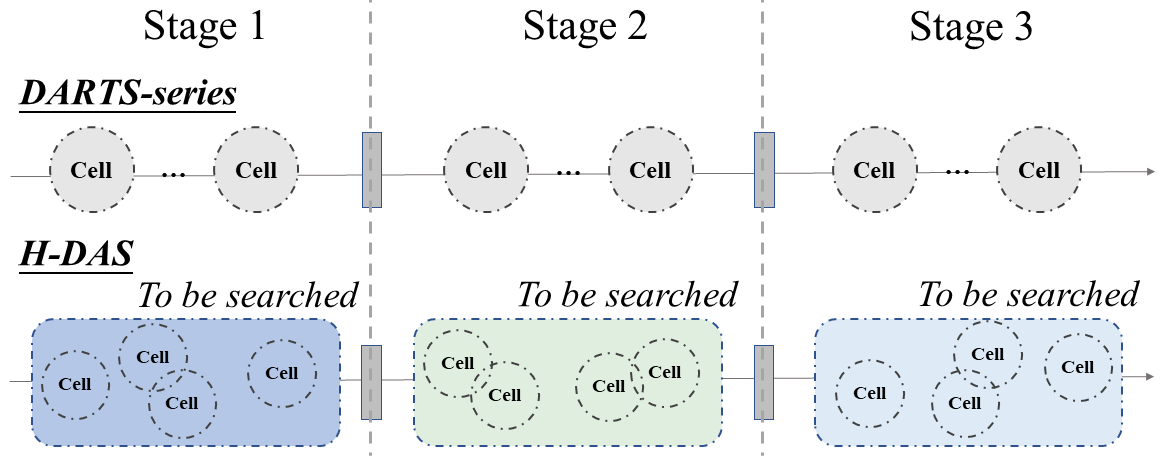}
  \caption{DARTS-series methods search for a repeatable cell structure, which is stacked sequentially in multiple stages. The proposed H-DAS enables a \textit{stage-specific} search of both cell-level and stage-level structures, allowing for a diversity of cell structures, cell distribution and cell connections  over different stages. This results in a significant larger search space with more meaningful architecture search which improves the performance.}
  \label{fig:teaser}
  \vspace{-5mm}
\end{figure}

Reinforcement learning~\cite{zoph2016neural,pham2018efficient} or evolutionary algorithm~\cite{real2019regularized,real2017large} has been introduced to NAS, due to the discrete nature of the architecture space. However, these methods usually require up to thousands of GPU days~\cite{zoph2016neural}.
A few methods have been developed to reduce the computational cost, such as ~\cite{cai2018proxylessnas,xie2018snas,dong2019searching,you2020greedynas,zhong2020representation}. Among them, differentiable architecture search~\cite{liu2018darts} attempted to approximate the discrete search space into a continuous one, where gradient descent can be used to optimize the architectures and model parameters jointly.
This line of search approaches, referred as DARTS-series methods~\cite{chen2019progressive,xu2019pc,chen2020stabilizing}, has made significant improvements in search speed, while maintaining comparable performance.

However, these DARTS-series methods have two major limitations in terms of search space. First, they commonly perform a cell-level search and adopt the same searched cell structure repeatedly for multiple stages (separated by the reduction cells), which may make the cell structure sub-optimal in the stages, since the optimal cell structures (including the connections and kernel sizes) at different stages can be significantly different. For example, in the architectures searched  by~\cite{zoph2016neural,cai2018proxylessnas}, the operations in shallow layers are mainly $3\times3$ convolutions, while many larger kernels appear in the deeper layers.
Second, previous DARTS-series methods mainly focus on searching for a repeatable cell structure, which is stacked sequentially to form the networks with three stages.
This configuration assumes a simple chain-like structure at the stage level, which reduces the search space considerably, and ignores  the importance of stage-level connections or structures.
As revealed by~\cite{yang2019evaluation}, the overall stage-level connections can impact to the final performance of the networks considerably.

In this work, we redesign the search space of DARTS-series methods, and propose a Hierarchical Differentiable Architecture Search (H-DAS) that enables the search of both cell-level (micro-architecture) and stage-level structures (macro-architecture) (Figure~\ref{fig:teaser}). H-DAS significantly increases the search space comparing to previous methods.
Specifically, for the micro-architecture, we relax the cell-level search space so that the networks can learn the optimized cell structures at different stages.
For searching the macro-architecture, we model each stage as a Directed Acyclic Graph (DAG), where each cell is a node of the DAG.

However, naively searching for the macro-architectures inevitably increases a large amount of additional parameters with corresponding computational overhead. More importantly, directly applying the method of cell search for searching the stages can lead to performance degradation, such as flattened stage structures. To address these issues, we carefully design three search rules and a depth loss, which allow us to systematically study the architectures at the stage level.
First, we propose a novel yet simple method to search for the distribution of cells over different stages, under a constraint of computational complexity. This allows for a better optimization on the numbers of cells, which is never investigated in previous DARTS-series methods, where all stages are manually set to have the same number of cells. 
Second, with the optimized cell distribution computed in the previous step, we then focus on the search of the stage-level architecture. 
To the best of our knowledge, we are, for the first time, to explore the stage-level  macro-architecture search, by relaxing the topological structures among different stages. We show that the proposed H-DAS can improve the performance for image classification prominently. The contributions of this work are summarized as follows:

- We propose a two-level Hierarchical Differentiable Architecture Search (H-DAS) that searches for structures at both the cell level and the stage level. 
H-DAS includes a cell-level search ($H^{c}$-DAS) and a stage-level search  ($H^{s}$-DAS), for the micro- and macro-architectures, respectively.

- $H^{c}$-DAS  is able to search for stage-specific cell structures within a greatly enlarged search space, comparing to that of DARTS. $H^{s}$-DAS includes a number of carefully-designed search rules and losses, which allows it to first search for an optimal distribution of cells over different stages, and then perform the search again to find the optimal structure for each stage.

- We conduct extensive experiments to demonstrate the effectiveness of the proposed H-DAS, which achieves a 2.41\% test error on CIFAR10 and a 24.5\% top-1 error on ImageNet. 
Moreover, the proposed stage-level search can be conducted based on the cell structures explored by other DAS methods, which further improves the performance.

\begin{figure*}[!t]
  \centering
  \includegraphics[width=1\textwidth]{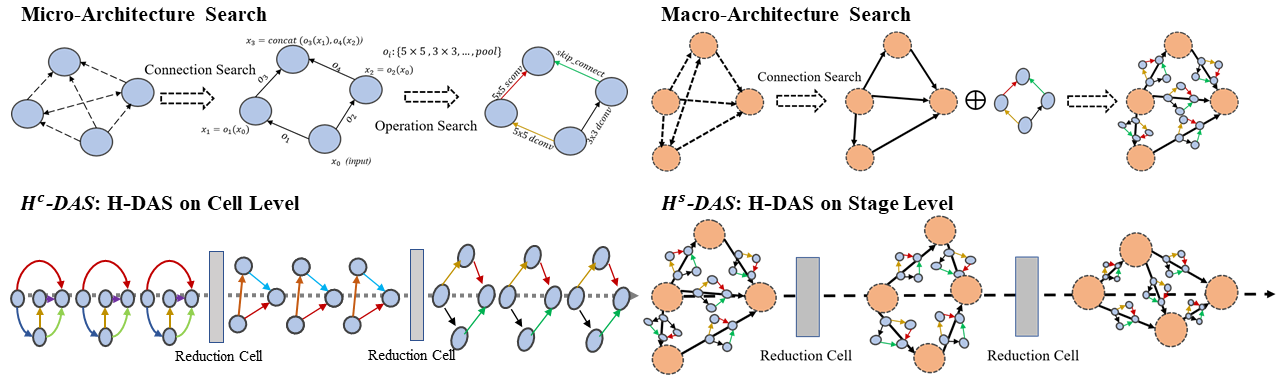}
  \caption{The overall pipeline of H-DAS, which includes a cell-level micro-architecture search (${H^{c}}$-DAS) and a stage-level macro-architecture search (${H^{s}}$-DAS) (best viewed in color).
   ${H^{c}}$-DAS searches for the structure of the cell, including operations (nodes) and connections between nodes. It relaxes the cell-level search space to learn \textit{stage-specific} cell structures over different stages.
 ${H^{s}}$-DAS searches for the connections and operations between the cells in the stage-level search space.} 
  \label{fig:main}
\end{figure*}

\section{Related Work}
Neural architecture search (NAS) has recently been attracting more attentions, and it can be defined as searching for an optimal operation out of a defined operation set and the  best connectivity between the operations, by using a Directed Acyclic Graph (DAG) ~\cite{zela2020bench}.
The weight-sharing paradigm led to a significant improvement in search efficiency.
Differentiable Architecture Search (DARTS)~\cite{liu2018darts} relaxed the discrete search space to be continuous, making it possible to search architectures and learn network weights using gradient descent. 
P-DARTS~\cite{chen2019progressive} focused on bridging the depth gap between a search stage and an evaluation stage. PC-DARTS~\cite{xu2019pc} performed a more efficient search without comprising the performance by sampling a small part of super-net to reduce the redundancy in a network space. In SmoothDARTS~\cite{chen2020stabilizing}, a perturbation-based regularization was proposed to smooth the loss landscape, and improve the generalizability. FairDARTS~\cite{chu2019fair} solved the problem of aggregation of skip connections.
These approaches follows the convention by  stacking identical cells to form a chain-like structure.

Our stage-level search is related to that of~\cite{liu2017hierarchical}, where low-level operations are assembled into a high-level motif, but the conceptions of states were not explored, and an evolutionary algorithm was applied for optimizing the search, making it much less efficient.
In ~\cite{liang2019computation}, a computation reallocation (CR) was developed to search for the stage length for object detection, which inspired the current work, but our approach is more efficient by designing a hierarchical search that performs both cell-level and stage-level search jointly for a different target task.

\section{Methodology}
\paragraph{Preliminary.}
In this work, we follow the cell-level design of DARTS~\cite{liu2018darts}. The goal of DARTS is to search for a repeatable cell, which can be stacked to form a convolutional network.
Each cell is a directed acyclic graph (DAG) of $N$ nodes $\{x_{i}\}_{i=0}^{N-1}$, where each node can be represented as a network layer. 
Weighted by the architecture parameter $\alpha^{(i,j)}$, each edge $(i,j)$ of DAG indicates an information flow from node $i$ to node $j$, and is formulated as:

\vspace{-3mm}
\begin{equation}
    f_{i,j}(\mathbf{x}_{i})=\sum_{o\in O_{i,j}}\frac{exp\{\alpha_{i,j}^o\}}{\sum_{o^{'}\in O}exp\{\alpha_{i,j}^{o^{'}}\}}o(\mathbf{x}_{i}).
\label{formula 1}
\end{equation}
where $\mathbf{x}_i$ is the feature map at the $i$-th node, and $o$ denotes candidate operations.
More details, such as bi-level optimization, can be found in~\cite{liu2018darts}.

\paragraph{Hierarchical search space.}
In this work, we redesign the search space of DARTS-series methods, and propose a Hierarchical Differentiable Architecture Search (H-DAS) that enables the search both at  the cell level (for micro-architecture) and at the stage level  (for macro-architecture).
As shown in figure~\ref{fig:main},  we relax the cell-level search space so that the network can learn the stage-specific cell structures for micro-architecture search. Then we model each stage as a DAG for macro-architecture search, which increases the variety of connections between cells. The two methods are named as $H^{c}$-DAS (for cell level) and $H^{s}$-DAS (for stage level), respectively.

\subsection{Micro-Architecture}
To enrich cell structures, we design $H^{c}$-DAS that relaxes the cell-level search space so that the networks can learn more meaningful stage-specific cell structures over different stages.
In an extreme case, one can search for a specific structure for each cell, which results in a maximum of cell-level search space.
However, it is not practical to performance NAS with such a large search space, which may make
the NAS search process not stable, because the search can be influenced by many factors like hyper-parameters, the competition between model weights, and architecture parameters during bi-level optimization~\cite{liang2019darts+}. 
More importantly, it is difficult to set the depth of networks flexibly for different goals when the network has a unique structure per cell. \\

\noindent
\textbf{Search space relaxation.}
To relax the cell-level search space, and maintain the stability during the search period, our $H^{c}$-DAS aims to search for \textit{stage-specific} cell-level structures, where the stages can have their own cell structures to capture different levels of semantics, while all cells in each stage has the same structure searched.
Similar to DARTS-series methods, we use two reduction cells to divide the spatial resolution by 2 at 1/3 and 2/3 of the total depth of the networks, which separate the entire networks into three stages. The cell search involves both connections search and operation search.
The goal of $H^{c}$-DAS is to find three structures of normal cells, each of which could be stacked repeatedly to form the optimal cell structure in the corresponding stage. The motivation of this design is that the shallow layers of CNNs often focus on learning low-level image information, like texture, while the deep layers of CNNs pay more attention on high-level  information, like semantic features. Therefore, the optimal cell structures at different stages should be diverse and play different functions, and our $H^{c}$-DAS naturally enriches the searched cell structures and increases the search space of the micro-architecture considerably.

\subsection{Macro-Architecture}
DARTS and its extensions mostly focus on searching for a repeatable cell structure, which is then stacked repeatedly and sequentially in a chain over multiple stages to form the networks.
This setting assumes a simple chain-like structure for multiple stages, which significantly reduces the search space, and ignores the diversity in stage-level structures.
Similarly, recent NAS methods based on mobile inverted bottleneck~\cite{sandler2018mobilenetv2}, such as MnasNet~\cite{tan2019mnasnet}, also stack the repeatable MBConvs sequentially to form the networks.

In addition to the search of cell structure, we found that it is important to build meaningful high-level macro-architecture by searching for the optimal connections between the cells over different stages. In this work, we introduce a Directed Acyclic Graph (DAG) structure to stage-level macro-architecture search. %
As shown in Figure~\ref{fig:main}, $H^{s}$-DAS searches for the connections between the searched (normal) cells in each stage to form a macro-architecture, which allows us  to explore the power of stage-level structures.
Notably, the searched macro-architectures can vary at different depths of networks, allowing the networks to learn meaningful cell connections at different stages.

\begin{figure}[t]
  \begin{center}
  \subfigure[$m=7$]{
  \label{3d.sub.1}
  \includegraphics[width=0.20\textwidth,trim=0 0 0 0, clip=true]{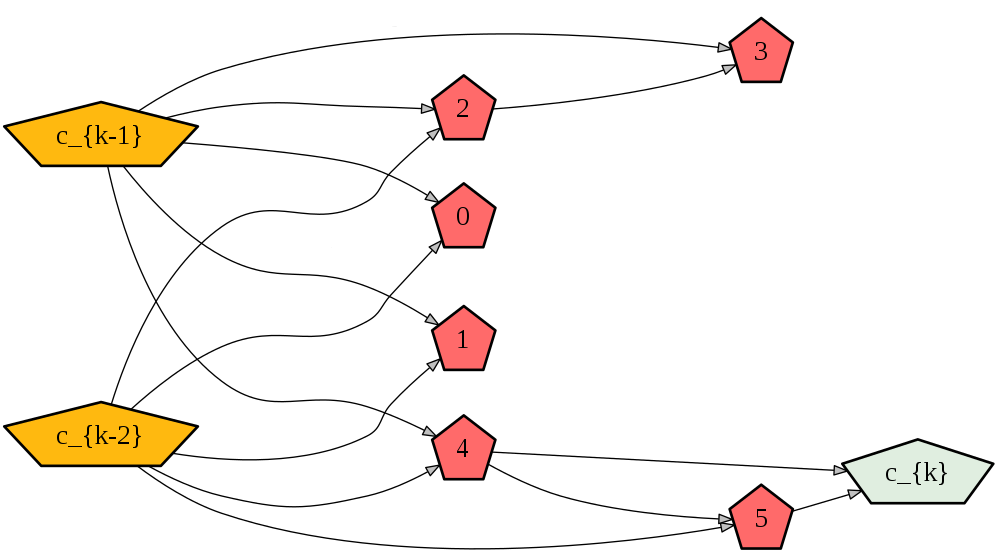}
  }
  \subfigure[$m=4$]{
  \label{3d.sub.2}
  \includegraphics[width=0.22\textwidth,trim=0 0 0 0, clip=true]{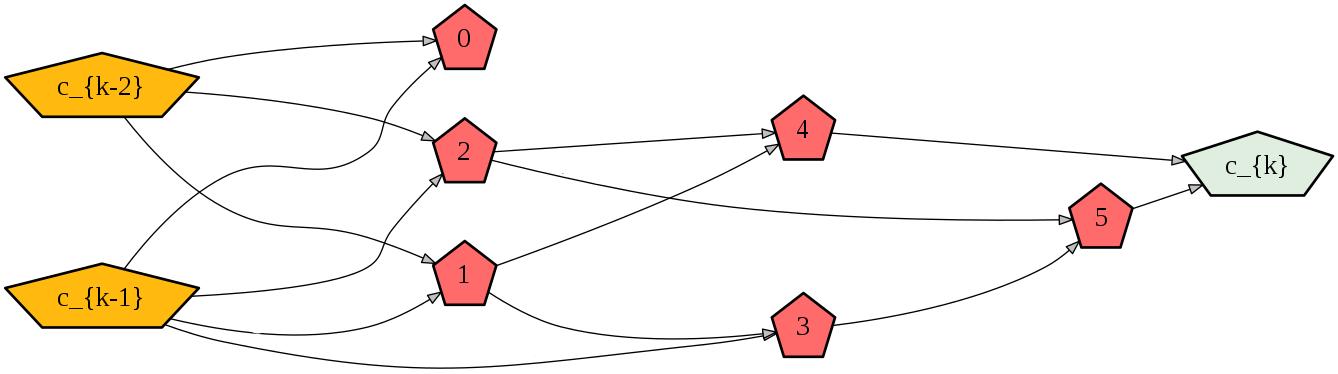}
  }
  \subfigure[stage-level structure ($m=3$)]{
  \label{3d.sub.3}
  \includegraphics[width=0.45\textwidth,trim=0 0 0 0, clip=true]{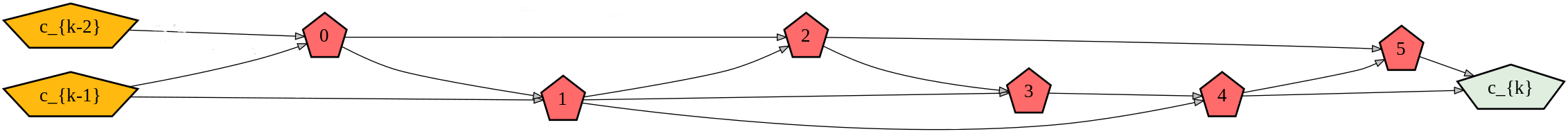}
  }
  
  \end{center}
    \caption{Comparison on the number of candidate input cells. The stage level outside structure with $N$ cells, and $m$ is the number of previous candidate cells which the current cell could connect with, like a sliding window of $m$ length, where $m\in[2, N+1]$.}
  \label{fig:change DAG}
\end{figure}

However,  searching for the macro-architecture is non-trivial, and would suffer from several problems. For example, many additional parameters and computational costs will be introduced by performing the search in the new stage-level search space. Moreover, the searched stage-level structures may become very shallow due to the ease on optimizing shallow networks, which will degrade the performance of networks. 
In this work, we carefully design three important rules, with a novel depth loss, to ensure a robust and efficient search of the stage-level structure. \\

\noindent\textbf{Rule 1: non-parametric connections. }
We design a set of candidate operations for the stage-level search space based on a key observation: the connections between cells may play a more important role than operation types.
This observation is demonstrated by ablation experiments as shown in supplementary material (SM), and it was also discussed in~\cite{xie2019exploring}.
Hence, we define a small set of candidate operations, including \textit{avg pooling}, \textit{max pooling}, \textit{skip-connect} and \textit{None}, which do not have any learnable parameter, and therefore keep the network capacity similar to the conventional sequential structure. 
The search process can be formulated as Eq. (\ref{formula 1}), with a difference that $\textbf{x}_i$ represents the output of the $i$-th cell.
In this case, the conventional stage-level structure of DARTS can be considered as a special case of $H^{s}$-DAS, when only a single operation \textit{skip-connect} is used between the cells. \\

\noindent\textbf{Rule 2: stage output with selective cell aggregation.}
In the cell-level search, the final output of an entire cell is a depth-wise concatenation of the outputs of all nodes within the cell. In this setting, all nodes can contribute to the output in terms of computation. However, in the stage-level search by $H^{s}$-DAS, when the same concatenation strategy is adopted, the channel size of the output of each stage can be significantly increased, which in turn results in a great increase of the parameters. 
To minimize the additional parameters introduced by $H^{s}$-DAS, we compute the output of a stage by using the concatenated features of the last two cells in the stage. 
However, this rule introduces another issue that some intermediate cells are not directly connected to any subsequent cells, and hence are not included in the computational graph, e.g. Figure~\ref{fig:change DAG} (a). We alleviate this problem by introducing Rule 3 as follows.\\

\noindent\textbf{Rule 3: constraint on preceding cells. }
A cell is defined as a dead cell when it is not connected to the stage output by any path in the graph. We empirically found that the performance of the networks is negatively impacted by the number of dead cells, since the complexity and  depth of the networks can be largely reduced when there are many dead cells, as shown in Figure \ref{fig:change DAG}.
To alleviate this problem, we set a constraint on the number of preceding cells which each current cell can connect to. 
Namely, the cells as candidate predecessors can be considered as a sliding window for each cell in the stage. For example, a cell can only choose its preceding cells from three previous cells when we set $m=3$ (Figure~\ref{fig:change DAG}).
This search rule can be considered as a trade-off between the stage-level search space and the number of active cells applied in the networks.  \\

\noindent\textbf{Depth loss. }
Without any restriction, the cells tend to directly connect to the input nodes, since shallower networks are generally easier to be optimized  during the search (with respect to the network parameters), but often have lower performance.
To alleviate this problem, we introduce a depth loss which takes into account the depth of the networks during the search.
Each cell in the stage has a depth number, which indicates the number of intermediate cells between the current cell and input feature maps. 
We set the depth number of the input feature maps to be 0.
In this case, the depth number of a cell is the weighted  sum (i.e. $\alpha$) of the depth numbers of its connected preceding cells.
Therefore, we can calculate the depth number for each cell in a recursive manner:

\begin{equation}
    L_{depth}=-\sum_{stage=1}^{3}\frac{1}N\sum_{i=0}^{N}\frac{1}i\sum_{j=0}^{i-1}\alpha_{i,j}\cdot(d_j+1),
\label{formula 2}
\end{equation}
where $\alpha_{i,j}$ indicates the weight between node $i$ and $j$, $d_j$ is the depth of node $j$, and $N$ is the number of cells in the stage. 
Minimizing the depth loss encourages the networks to go deep, and therefore improves the capability of networks. \\

\begin{figure}[t]
  \begin{center}
  \subfigure[Over-parameterized structure and candidate cells pairs]{
  \label{3d3c.bd.1}
  \includegraphics[width=0.42\textwidth]{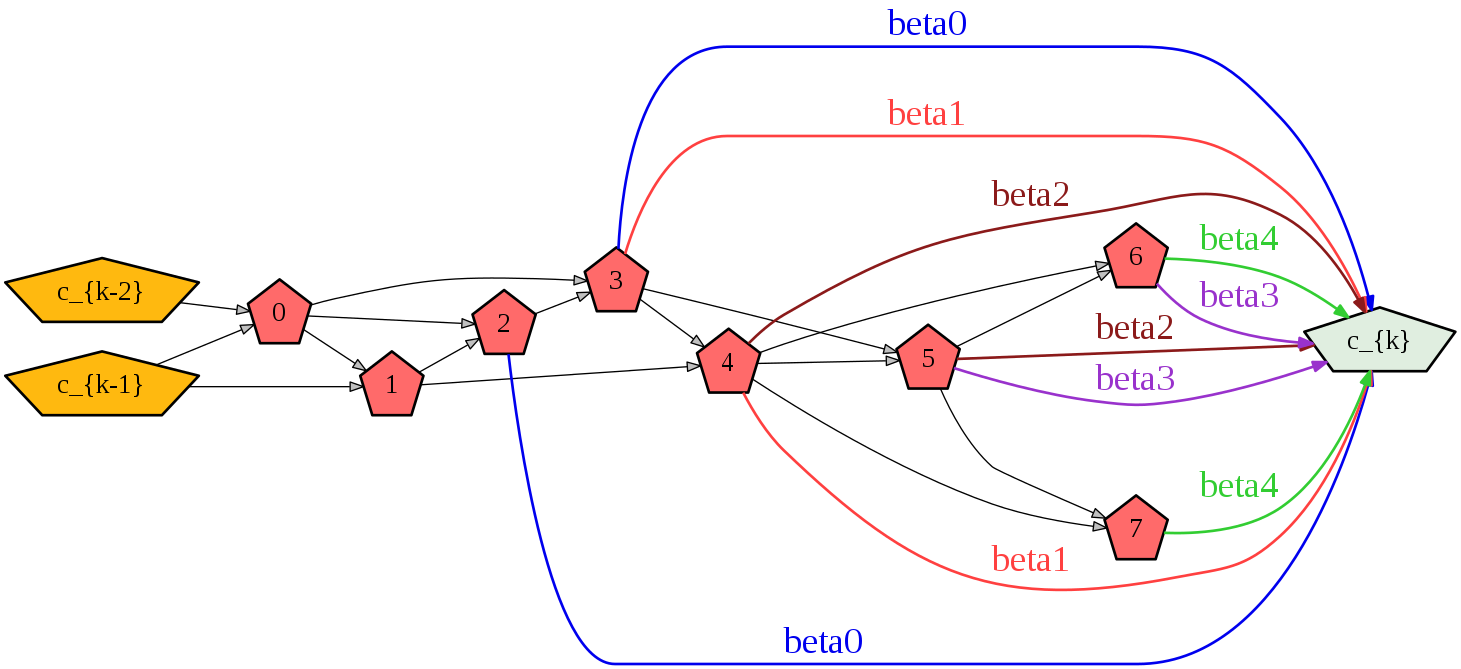}
  }
  \hfill
  \subfigure[Selection of an output-connected pair of cells]{
  \label{3d3c.bd.2}
  \includegraphics[width=0.35\textwidth]{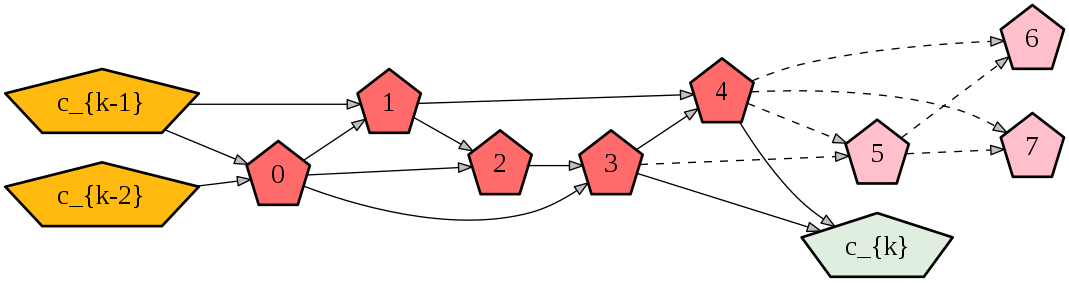}
  }
  
  \end{center}
    \caption{Search for distribution of cells in each stage by $\beta$.}
  \label{fig:changestage}
\end{figure}

\noindent\textbf{Search for distribution of cells.  }
With the three search rules and the new depth loss, we can now systematically study our stage-level architecture. First, we search for the number of cells in each stage.
Previous DARTS-series methods mainly follow conventional configuration by manually setting the same number of cells for all stages, and it has not been verified whether such manual configuration is optimal.
We therefore develop a simple method to explore the distribution of cells over different stages.
The key idea of this search is to initialize an over-parameterized network (i.e. by containing more cells than necessary), and then remove the less impactful cells in each stage during the search, under a constraint of certain network capacity or FLOPs.
Based on $Rule 2$, we introduce a parameter $\beta$ to encode the importance of five pairs of adjacent cells that can connect to the output cell, among which only one pair will be selected at the end of the search. The output of a stage can be formulated as:

\begin{equation}
  Output_{stage}=\sum softmax(\beta_{i,j})\cdot concat(X_i, X_j)
\end{equation}
where $\beta_{i,j}$ is the weight of output from cell $i$ and cell $j$, and $X_i$ are the output features of cell $i$. $\beta$ is an architecture parameter that selects the one optimal pair of cells connecting to the final output of each stage, 
as illustrated in Figure \ref{fig:changestage}.

\noindent To incorporate $\beta$, E.q. \ref{formula 2} can be reformulated as:

\begin{equation}
  \begin{split}
    L_{depth}=&-\sum_{stage=1}^{3}\frac{1}N\sum_{i=0}^{N}\frac{1}i\sum_{j=0}^{i-1}\alpha_{i,j} \\ 
              &\times(d_j+1)\times(1 + i\times softmax(\beta_{i-2})).
  \end{split}
\end{equation}

\noindent Crucially, a loss is adopted to constrain the computational complexity of the whole networks, and therefore the less important cells can be removed during the search. The cells in each stage share the same cell-level structure, and thus the computational complexity is constant for each stage. 
As a result, the loss with a constraint on the computational complexity can be simplified as follows:

\begin{equation}
    L_{comp}=\sum_{i=0}^{S}\theta_i\sum_{j=0}^{N-N_{min}}softmax(\beta_{i,j})\cdot(j+N_{min}),
\label{eq:fmin}
\end{equation}

where $N$ is the number of cells in each stage, and $N_{min}$ is the minimum number of cells in a stage.
$\theta_i$ can be either multi-adds (FLOPs) or the number of parameters, depending on the desired constraint.
We choose to have a constraint on the FLOPs. In this case, $\theta_i=1$ since the stages share the same cell structure and thus the FLOPs of the cell are the same across three stages.
The final loss function is:

\begin{equation}
    Loss = L_{cls} + \delta L_{depth} + \gamma\times L_{comp},
\label{f5}
\end{equation}
where $\delta$ and $\gamma$ are weighting factors that balance the contributions of different losses. 
After training, we choose the pair of cells which has the largest $\beta$ to connect to the output feature map.
By setting the weights in E.q.~\ref{f5}, we can explore the number of cells in each stage, and obtain a network with a configurable total number of cells.

Notably, the weight of depth loss $\delta$ influences the final performance of the searched architecture. The optimal weight can be explored empirically.
The corresponding experimental results regarding $\delta$ could be found in SM. \\

\noindent\textbf{Search for stage-level structure.}
We fix the number of cells in each stage when searched, and then focus on searching for a stage-level architecture.
The whole networks are constructed in the same manner as DARTS when the search is done.
We can directly derive the network structure in the search of cells, but we empirically found  that searching again with a fixed number of cells in each stage can improve the performance. This may be explained by a tighter search space which makes the search easier.
With the relaxation of the topological structures among stages, we have a hierarchical structure that contains the search of nodes, cells and stages.
The searched stage-level structure by $H^s$-DAS can be combined with other cell-level structures searched by existing DAS methods to further boost the performance. \\

\noindent\textbf{Search space complexity.  }
The proposed methods significantly increase the scale of search space. Specifically, $H^c$-DAS increases the search space of DARTS from $10^{18}$ to $10^{45}$ by allowing for stage-specific cells. Furthermore, by unchaining the conventional macro-architecture and searching for a specific macro-architecture for each stage, $H^s$-DAS has a search space of $10^{24}$ for a 20-cell network, without considering the cell-level search space. 
In the search of cell distribution by $H^s$-DAS, two additional cells are added in each stage, increasing the scale of the search space to $10^{33}$, which is considerably larger than those in previous DAS methods. 
By including the cell-level search space, the full search space of a 20-cell $H$-DAS can reach to $10^{42}$, which is significantly larger than previous methods, and naturally leads to a higher performance.
The details of the complexity calculation are described in SM.

\begin{table*}[t]
    \small
    \begin{center}
    \begin{tabular*}{0.9\hsize}{@{}@{\extracolsep{\fill}}lcccc@{}}
    \hline
    \multirow{2}{*}{\textbf{Architecture}}              & \textbf{Test Err.} & \textbf{Params} & \textbf{Search Cost} & {\textbf{Search}} \\
                                                        & \textbf{(\%)}      & \textbf{(M)}    & \textbf{(GPU-days)}  & {\textbf{Method}}                      \\ \shline
    ResNet~\shortcite{he2016deep}                    & 4.61               & 1.7             & -                    & manual                                  \\
    DenseNet-BC~\shortcite{huang2017densely}                  & 3.46               & 25.6            & -                    & manual                                  \\ \hline
    NASNet-A~\shortcite{zoph2018learning}            & 2.65               & 3.3             & 2000                 & RL                                      \\
    AmoebaNet-A~\shortcite{real2019regularized}      & 3.34               & 3.2             & 3150                 & evolution                               \\
    Hierarchical evolution~\shortcite{liu2017hierarchical}    & 3.75               & 15.7            & 300                  & evolution                               \\
    ENAS~\shortcite{pham2018efficient}               & 2.89               & 4.6             & 0.5                  & RL                                      \\ 
    Arch2Vec~\shortcite{yan2020does}                 & 2.56               & 3.6            & 100            & BO   \\
    \hline
    ProxylessNAS~\shortcite{cai2018proxylessnas}$^\dagger$              & 2.08               & 5.7               & 4                    & gradient                          \\
    DARTS(2nd order)~\shortcite{liu2018darts}                 & 2.76               & 3.3             & 1                    & gradient                          \\
    SNAS (mild)~\shortcite{xie2018snas}                         & 2.98               & 2.9             & 1.5                  & gradient                          \\
    GDAS(FRC)~\shortcite{dong2019searching}                   & 2.82               & 2.5             & 0.17                 & gradient                          \\
    P-DARTS~\shortcite{chen2019progressive}          & 2.62$^\star$ / 2.50                & 3.4             & 0.3                  & gradient                          \\
    PC-DARTS~\shortcite{xu2019pc}                    & 2.57               & 3.6             & 0.1                  & gradient                          \\
    NoisyDARTS~\shortcite{chu2020noisy}                    & 2.65$^\star$ / 2.39               & 3.6             & 0.4                 & gradient \\
    RDARTS ~\shortcite{zela2019understanding}                   & 2.95               & -             & 1.6                 & gradient \\
    SDARTS-ADV~\shortcite{chen2020stabilizing}              & 2.61               & 3.3             & 1.3                  & gradient                           \\
    FairDARTS~\shortcite{chu2019fair}                       & 2.54               & 3.3             & 0.41                  &  gradient                          \\
    ISTA-NAS~\shortcite{yang2020ista}                       & 2.54               & 3.3             & 0.05                 & gradient                          \\           
    \hline
    $H^{c}$-DAS                               & 2.66               & 2.3             & 0.4                  & gradient                         \\
    $H^{s}$-DAS                               & 2.41               & 3.4             & 0.7$^\diamond$                 & gradient                          \\
    $H^{s}$-DAS (with cell in P-DARTS)                              & 2.30               & 3.5             & 0.3                  & gradient                          \\
    $H^{s}$-DAS (with cell in NoisyDARTS)                              & 2.34               & 3.6             & 0.3                  & gradient                          \\
    $H^{s}$-DAS-autoAugment~\shortcite{cubuk2018autoaugment}                   & 1.99               & 3.6             & 0.7$^\diamond$                 &          gradient  \\   \hline
    \end{tabular*}
    \end{center}
    \caption{Comparison with state-of-the-art architecture on CIFAR10. ~ $^{\star}$: Our implementation by training the best cell architecture provided by the authors using the code of H-DAS. ~ $^\dagger$: Obtained on a different search space with PyramidNet~\cite{han2017deep} as the backbone. ~$^\diamond$: The search cost contains 0.4 GPU-day for cells and 0.3 GPU-day for stages.}  
    \label{tab:cifar10}
  \end{table*}

\section{Experiments and Results}
\subsection{Implementation Details}
We conduct experiments on CIFAR10~\cite{krizhevsky2009learning} and ImageNet~\cite{deng2009imagenet}.
In the search of cell-level structure, we follow DARTS~\cite{liu2018darts} by using the same search space, hyperparameters and training scheme.
We set $N_{min}=4$ in Eq.~(\ref{eq:fmin}).
The stage-level search space contains 4 non-parametric operations which connect the cells, including: \textit{$3\times3$ average pooling}, \textit{skip connection}, \textit{$3\times3$ max pooling}, \textit{no connection (none)}.
For training a single model, we use the same strategy and data processing methods as DARTS. More details can be found in SM.

\begin{table*}[t]
  \begin{center}
  \small
  \begin{tabular*}{0.95\hsize}{@{}@{\extracolsep{\fill}}lcccccc@{}}
  \hline
  \multirow{2}{*}{\textbf{Architecture}} & \multicolumn{2}{c}{\textbf{Test Err.(\%)}} & \textbf{Params} & \textbf{$\times+$} & \textbf{Search Cost} & {\textbf{Search}} \\ \cline{2-3}
                                         & \textbf{top-1}       & \textbf{top-5}      & \textbf{(M)}    & \textbf{(M)}       & \textbf{(GPU-days)}  & \textbf{Method}                                        \\ \shline
  Inception-v1~\shortcite{szegedy2015going}         & 30.2            & 10.1            & 6.6     & 1448      & -           & manual                         \\
  MobileNet-v2~\shortcite{sandler2018mobilenetv2}        & 25.3            & -            & 6.9     & 585       & -           & manual                         \\
  ShuffleNet 2x (v2)~\shortcite{ma2018shufflenet}    & 25.1            & -               & 7.4 & 591       & -           & manual                         \\ \hline
  NASNet-A~\shortcite{zoph2018learning} $^{\dagger\dagger}$            & 26.0            & 8.4             & 5.3     & 564       & 1800        & RL                             \\
  AmoebaNet-C~\shortcite{real2019regularized} $^{\dagger\dagger}$      & 24.3            & 7.6             & 6.4     & 570       & 3150        & evolution                      \\
  PNAS~\shortcite{liu2018progressive} $^{\dagger\dagger}$              & 25.8            & 8.1             & 5.1     & 588       & 225         & SMBO                           \\
  MnasNet-92~\shortcite{tan2019mnasnet} $^{\dagger\dagger}$            & 25.2            & 8.0             & 4.4     & 388       & -           & RL                             \\ 
  MobileNet-v3-large~\shortcite{howard2019searching}$^{\dagger\dagger}$ & 24.8           & -               & 5.4     & 219       & -           & RL                             \\ 
  \hline
  DARTS (2nd order)~\shortcite{liu2018darts}  $^\dagger$       & 26.7            & 8.7             & 4.7     & 574       & 1           & gradient                 \\
  GDAS~\shortcite{dong2019searching} $^\dagger$               & 26.0            & 8.5             & 5.3     & 581       & 0.21        & gradient                 \\
  SNAS (mild)~\shortcite{xie2018snas}   $^\dagger$             & 27.3            & 9.2             & 4.3     & 522       & 1.5         & gradient                 \\
  P-DARTS~\shortcite{chen2019progressive}$^\dagger$  & 25.1$^\star$ / 24.4            &7.7$^\star$ / 7.4             & 4.9     & 557       & 0.3         & gradient                 \\
  SinglePath-NAS~\shortcite{stamoulis2019single}$^{\dagger\dagger}$    & 25.0            & 7.8              & -      & -         & 0.15        & gradient                 \\
  ProxylessNAS (GPU)~\shortcite{cai2018proxylessnas}$^{\dagger\dagger}$ & 24.9           & 7.5              & 7.1    & 465       & 8.3         & gradient                 \\
  PC-DARTS~\shortcite{xu2019pc}  $^\dagger$          & 25.1            & 7.8             & 5.3     & 586       & 0.1         & gradient                 \\
  RandWire-WS~\shortcite{xie2019exploring} $^{\dagger\dagger}$         & 25.3            & 7.8             & 5.6     & 583       & -           & random                      \\ 
  SDARTS-ADV~\shortcite{chen2020stabilizing} $^\dagger$       & 25.2            & 7.8             &  -         & -         & 1.3         & gradient  \\
  FairDARTS~\shortcite{chu2019fair}  $^\dagger$       & 24.9         & 7.5               & 4.8     & 541       & 0.4           &      gradient                \\   
  ISTA-NAS~\shortcite{yang2020ista} $^\dagger$        & 25.1        & 7.7           & 4.78     & 550    & 2.3          & gradient  \\
  \hline
  $H^{c}$-DAS $^\dagger$                     & 25.9            & 8.4               & 5.0     & 578       & 0.5         & gradient                \\
  $H^{s}$-DAS   $^\dagger$                   & 24.5            & 7.7               & 5.1     & 572       & 0.3        & gradient                 \\ \hline
  \end{tabular*}
  \caption{Comparison with state-of-the-art architectures on ImageNet (mobile setting). ~ $^{\star}$: Our implementation by training the best architecture provided by the authors using the code of H-DAS. ~ $^\dagger$: Searched on CIFAR10. ~ $^{\dagger\dagger}$: Searched on ImageNet.}
  \label{tab:imgnet}
  \end{center}
\end{table*}

\subsection{Search for Cell Distribution}
The search of cell distribution over three stages is performed  under a constraint of  certain computational complexity. 
It is interesting that different stages can have different numbers of cells at the beginning of the search or during the search, but the numbers of cells in all stages become the same at the end of the search.
As discussed, adjusting the weighting factor $\gamma$ in Eq.~(\ref{f5}) leads to a different total number of cells in the networks. 
We therefore repeated the search several times with various values of $\gamma$, resulting in a different total number of cells in the networks. But the cell distribution remains the same.
Based on these observations, for a fair comparison, we set the number of cells in each stage in our $H^s$-DAS to be 6 for CIFAR10 and 4 for ImageNet, which are the same as other DARTS-series methods. 

\subsection{Results on CIFAR10}
The results on CIFAR10 are compared in Table~\ref{tab:cifar10}. 
$H^{c}$-DAS has many non-parametric connections in the cells of the last stage, and thus its parameter size becomes  small, which is about 30\% smaller than that of DARTS.
However, it still has better performance than DARTS, suggesting that enlarging the cell-level search space and learning stage-specific cell-level structures could bring a great improvement. It also shows that searching for a single repeatable cell structure to form a chain-like network is not fully optimized.

Furthermore, by combining with the cell structures searched by recent excellent DAS methods~\cite{chen2019progressive,chu2020noisy}, our $H^s$-DAS, which searches for more meaningful cell connections over different stages, can improve the stage-of-the-art methods by achieving an error of 2.30\% on CIFAR10, which is the best among DAS methods.
Notice that ProxylessNAS has a better result, but with 60-100\% more parameters and $\times10$ search time comparing to other methods, including our $H^s$-DAS.

We also found that the performance of $H^{s}$-DAS is better than $H^{c}$-DAS, suggesting that the macro-architecture plays an important role in neural architecture. 
A simple chain-like sequential structure would significantly limit the search space, and discard the importance of stage-level structures. It is beneficial to unchain the search space and pay more attention on developing more meaningful macro-architectures in the future research.

\subsection{Results on ImageNet}
To verify the generalization of our searched structures, we train our networks on ImageNet using the architecture searched from CIFAR10. Results of our $H^{c}$-DAS and $H^{s}$-DAS on ImageNet are compared in Table~\ref{tab:imgnet}. We transfer the cell-level and stage-level structures with four normal cells searched from CIFAR10 to ImageNet.
Again, $H^{s}$-DAS outperforms $H^{c}$-DAS with a large margin, demonstrating the superiority of meaningful cell connections in stage-level structures.
Additionally, our $H^{s}$-DAS is comparable against other state-of-the-art DAS methods, without using the advanced network block design like MBConvs.

\begin{table}[t]
  \small
  \begin{center}
  \begin{tabular}{c|cccccc}
  \hline
  Search              & \multicolumn{2}{c|}{Cell} & \multicolumn{2}{c|}{Stage} & \multicolumn{2}{c}{Test Err.(\%)} \\ \cline{2-7} 
  Space               & Sh.        & Spec.       & Sh.        & Spec.        & C10            & ImgNet           \\ \shline
  DARTS               &  \cmark            &  \xmark          &   \textbf{-}          &   \textbf{-}          & 2.76           & 26.70            \\
  $H^c$-DAS           &  \xmark            &   \cmark         &    \textbf{-}        &   \textbf{-}        & 2.66           & 25.91            \\
  $H^s$-DAS$^\star$   &  \cmark            &   \xmark         &    \cmark          &    \xmark         & 2.58           & 25.45            \\
  $H^s$-DAS$^\dagger$ &   \xmark          &   \cmark         &      \xmark        &     \cmark        & 2.51           & 25.80            \\
  $H^s$-DAS           &    \cmark          &     \xmark       &      \xmark        &      \cmark       & 2.30           & 24.50            \\ \hline
  \end{tabular}
  \end{center}
  \caption{Performance of the cell-level search and stage-level search, by using the \textit{shared} structures over all stages or the \textit{stage-specific} architectures for different stages. 
  DARTS and  $H^c$-DAS do not perform stage-level search. }
  \label{tab:abla}
 \end{table}

\begin{table}[t]
    \small
    \begin{center}
    \begin{tabular}{ccccc}
    
      \hline
      Structure & m & \# Dead cells & \# Depth & Test Err.(\%) \\ \shline
      Fig.\ref{3d.sub.1}    & 7     & 4               & 2        & 3.08    \\
      Fig.\ref{3d.sub.2}    & 4     & 1               & 3        & 2.91    \\
      Fig.\ref{3d.sub.3}    & 3     & 0               & 5        & 2.58    \\ \hline
    \end{tabular}
    \end{center}
    \caption{Constraint of input cells. The three macro-architectures are retrained on CIFAR10 with the same configuration. \emph{'m'}: the length of sliding window. \emph{`\# Dead cells'}: the number of cells which are not connected to a subsequent cell. \emph{`\# Depth'}: the number of cells in the longest path from input to output.}
    \label{tab:abla2}

\end{table}

\subsection{Ablation Study}
To demonstrate the effectiveness of cell-level search and stage-level search, we do ablation experiments on CIFAR10 and ImageNet with various configurations. The details are reported in Table \ref{tab:abla}. 
We can summary that, (1) our $H^c$-DAS outperforms DARTS by allowing for searching for different cell structures over different stages;
(2) comparing to DARTS, our $H^s$-DAS$^\star$ which performs the stage-level search, can improve the performance of the unsearched chain-like structure;
(3)  $H^s$-DAS  can further improve $H^s$-DAS$^\star$ by implementing the \textit{stage-specific} macro-architecture search;
(4) it is interesting that $H^s$-DAS$^\dagger$ has a lower performance than $H^s$-DAS by performing the \textit{stage-specific} cell search. This would result in a larger search space which might be over-complex for a search algorithm to learn an optimal result.

Additionally, as we found empirically in Section 3.2 that the number of dead cells in the searched architecture would impact the performance of the networks.
We further verify it by training the rest two structures in Figure \ref{fig:change DAG} on CIFAR10. As presented in Table \ref{tab:abla2}, a network with less dead cells and a deeper macro-architecture obtains a higher performance.

Furthermore, to show the effectiveness of our searching strategy on macro-architecture, we keep the same cell-level structures and compare the macro-architectures searched with a random baseline.
By following~\cite{yang2019evaluation}, we calculate a relative improvement over this random baseline as $RI=100\times({Acc}_{m}-{Acc}_{r})/{Acc}_{r}$, which provides a quality measurement of the search strategy.
${Acc}_m$ and ${Acc}_{r}$ indicate the top-1 accuracy of a search method and the random sampling strategy, respectively. 
Our  $H^s$-DAS achieves a $RI$ of 0.44, with a significant improvement over 0.32 of DARTS.

\section{Conclusions}
We have presented our Hierarchical Differentiable Architecture Search (H-DAS) which preforms both cell-level and stage-level architecture search. 
In the cell-level search, H-DAS improves DARTS baseline by exploring stage-specific cell structures. Importantly, we formulate the stage-level structure as a directed acyclic graph, which allows us to search for a more advanced architecture than the conventional chain-like configuration.
Our two-level search enlarges the search space considerably, and improves the performance of previous DARTS-series methods significantly.

\bibliography{egbib}

\begin{thebibliography}{43}
\providecommand{\natexlab}[1]{#1}
\providecommand{\url}[1]{\texttt{#1}}
\providecommand{\urlprefix}{URL }
\expandafter\ifx\csname urlstyle\endcsname\relax
  \providecommand{\doi}[1]{doi:\discretionary{}{}{}#1}\else
  \providecommand{\doi}{doi:\discretionary{}{}{}\begingroup
  \urlstyle{rm}\Url}\fi

\bibitem[{Baker et~al.(2016)Baker, Gupta, Naik, and
  Raskar}]{baker2016designing}
Baker, B.; Gupta, O.; Naik, N.; and Raskar, R. 2016.
\newblock Designing neural network architectures using reinforcement learning.
\newblock \emph{arXiv preprint arXiv:1611.02167} .

\bibitem[{Cai, Zhu, and Han(2018)}]{cai2018proxylessnas}
Cai, H.; Zhu, L.; and Han, S. 2018.
\newblock Proxylessnas: Direct neural architecture search on target task and
  hardware.
\newblock \emph{arXiv preprint arXiv:1812.00332} .

\bibitem[{Chen and Hsieh(2020)}]{chen2020stabilizing}
Chen, X.; and Hsieh, C.-J. 2020.
\newblock Stabilizing Differentiable Architecture Search via Perturbation-based
  Regularization.
\newblock \emph{ICML} .

\bibitem[{Chen et~al.(2019)Chen, Xie, Wu, and Tian}]{chen2019progressive}
Chen, X.; Xie, L.; Wu, J.; and Tian, Q. 2019.
\newblock Progressive Differentiable Architecture Search: Bridging the Depth
  Gap between Search and Evaluation.
\newblock \emph{arXiv preprint arXiv:1904.12760} .

\bibitem[{Chu, Zhang, and Li(2020)}]{chu2020noisy}
Chu, X.; Zhang, B.; and Li, X. 2020.
\newblock Noisy Differentiable Architecture Search.
\newblock \emph{arXiv preprint arXiv:2005.03566} .

\bibitem[{Chu et~al.(2020)Chu, Zhou, Zhang, and Li}]{chu2019fair}
Chu, X.; Zhou, T.; Zhang, B.; and Li, J. 2020.
\newblock Fair darts: Eliminating unfair advantages in differentiable
  architecture search.
\newblock \emph{ECCV} .

\bibitem[{Cubuk et~al.(2018)Cubuk, Zoph, Mane, Vasudevan, and
  Le}]{cubuk2018autoaugment}
Cubuk, E.~D.; Zoph, B.; Mane, D.; Vasudevan, V.; and Le, Q.~V. 2018.
\newblock Autoaugment: Learning augmentation policies from data.
\newblock \emph{arXiv preprint arXiv:1805.09501} .

\bibitem[{Deng et~al.(2009)Deng, Dong, Socher, Li, Li, and
  Fei-Fei}]{deng2009imagenet}
Deng, J.; Dong, W.; Socher, R.; Li, L.-J.; Li, K.; and Fei-Fei, L. 2009.
\newblock Imagenet: A large-scale hierarchical image database.
\newblock In \emph{CVPR}.

\bibitem[{Dong and Yang(2019)}]{dong2019searching}
Dong, X.; and Yang, Y. 2019.
\newblock Searching for a robust neural architecture in four gpu hours.
\newblock In \emph{CVPR}.

\bibitem[{Han, Kim, and Kim(2017)}]{han2017deep}
Han, D.; Kim, J.; and Kim, J. 2017.
\newblock Deep pyramidal residual networks.
\newblock In \emph{Proceedings of the IEEE conference on computer vision and
  pattern recognition}, 5927--5935.

\bibitem[{He et~al.(2016)He, Zhang, Ren, and Sun}]{he2016deep}
He, K.; Zhang, X.; Ren, S.; and Sun, J. 2016.
\newblock Deep residual learning for image recognition.
\newblock In \emph{CVPR}.

\bibitem[{Howard et~al.(2019)Howard, Sandler, Chu, Chen, Chen, Tan, Wang, Zhu,
  Pang, Vasudevan et~al.}]{howard2019searching}
Howard, A.; Sandler, M.; Chu, G.; Chen, L.-C.; Chen, B.; Tan, M.; Wang, W.;
  Zhu, Y.; Pang, R.; Vasudevan, V.; et~al. 2019.
\newblock Searching for mobilenetv3.
\newblock In \emph{ICCV}.

\bibitem[{Huang et~al.(2017)Huang, Liu, Van Der~Maaten, and
  Weinberger}]{huang2017densely}
Huang, G.; Liu, Z.; Van Der~Maaten, L.; and Weinberger, K.~Q. 2017.
\newblock Densely connected convolutional networks.
\newblock In \emph{CVPR}.

\bibitem[{Krizhevsky, Hinton et~al.(2009)}]{krizhevsky2009learning}
Krizhevsky, A.; Hinton, G.; et~al. 2009.
\newblock Learning multiple layers of features from tiny images.
\newblock Technical report, Citeseer.

\bibitem[{Krizhevsky, Sutskever, and Hinton(2012)}]{krizhevsky2012imagenet}
Krizhevsky, A.; Sutskever, I.; and Hinton, G.~E. 2012.
\newblock Imagenet classification with deep convolutional neural networks.
\newblock In \emph{NIPS}.

\bibitem[{Liang et~al.(2019{\natexlab{a}})Liang, Lin, Guo, Sun, Wu, Yan, and
  Ouyang}]{liang2019computation}
Liang, F.; Lin, C.; Guo, R.; Sun, M.; Wu, W.; Yan, J.; and Ouyang, W.
  2019{\natexlab{a}}.
\newblock Computation reallocation for object detection.
\newblock \emph{arXiv preprint arXiv:1912.11234} .

\bibitem[{Liang et~al.(2019{\natexlab{b}})Liang, Zhang, Sun, He, Huang, Zhuang,
  and Li}]{liang2019darts+}
Liang, H.; Zhang, S.; Sun, J.; He, X.; Huang, W.; Zhuang, K.; and Li, Z.
  2019{\natexlab{b}}.
\newblock Darts+: Improved differentiable architecture search with early
  stopping.
\newblock \emph{arXiv preprint arXiv:1909.06035} .

\bibitem[{Liu et~al.(2018)Liu, Zoph, Neumann, Shlens, Hua, Li, Fei-Fei, Yuille,
  Huang, and Murphy}]{liu2018progressive}
Liu, C.; Zoph, B.; Neumann, M.; Shlens, J.; Hua, W.; Li, L.-J.; Fei-Fei, L.;
  Yuille, A.; Huang, J.; and Murphy, K. 2018.
\newblock Progressive neural architecture search.
\newblock In \emph{ECCV}.

\bibitem[{Liu et~al.(2017)Liu, Simonyan, Vinyals, Fernando, and
  Kavukcuoglu}]{liu2017hierarchical}
Liu, H.; Simonyan, K.; Vinyals, O.; Fernando, C.; and Kavukcuoglu, K. 2017.
\newblock Hierarchical representations for efficient architecture search.
\newblock \emph{arXiv preprint arXiv:1711.00436} .

\bibitem[{Liu, Simonyan, and Yang(2018)}]{liu2018darts}
Liu, H.; Simonyan, K.; and Yang, Y. 2018.
\newblock Darts: Differentiable architecture search.
\newblock \emph{arXiv preprint arXiv:1806.09055} .

\bibitem[{Ma et~al.(2018)Ma, Zhang, Zheng, and Sun}]{ma2018shufflenet}
Ma, N.; Zhang, X.; Zheng, H.-T.; and Sun, J. 2018.
\newblock Shufflenet v2: Practical guidelines for efficient cnn architecture
  design.
\newblock In \emph{ECCV}.

\bibitem[{Pham et~al.(2018)Pham, Guan, Zoph, Le, and Dean}]{pham2018efficient}
Pham, H.; Guan, M.~Y.; Zoph, B.; Le, Q.~V.; and Dean, J. 2018.
\newblock Efficient neural architecture search via parameter sharing.
\newblock \emph{arXiv preprint arXiv:1802.03268} .

\bibitem[{Real et~al.(2019)Real, Aggarwal, Huang, and Le}]{real2019regularized}
Real, E.; Aggarwal, A.; Huang, Y.; and Le, Q.~V. 2019.
\newblock Regularized evolution for image classifier architecture search.
\newblock In \emph{AAAI}.

\bibitem[{Real et~al.(2017)Real, Moore, Selle, Saxena, Suematsu, Tan, Le, and
  Kurakin}]{real2017large}
Real, E.; Moore, S.; Selle, A.; Saxena, S.; Suematsu, Y.~L.; Tan, J.; Le,
  Q.~V.; and Kurakin, A. 2017.
\newblock Large-scale evolution of image classifiers.
\newblock In \emph{ICML}.

\bibitem[{Sandler et~al.(2018)Sandler, Howard, Zhu, Zhmoginov, and
  Chen}]{sandler2018mobilenetv2}
Sandler, M.; Howard, A.; Zhu, M.; Zhmoginov, A.; and Chen, L.-C. 2018.
\newblock Mobilenetv2: Inverted residuals and linear bottlenecks.
\newblock In \emph{CVPR}.

\bibitem[{Simonyan and Zisserman(2014)}]{simonyan2014very}
Simonyan, K.; and Zisserman, A. 2014.
\newblock Very deep convolutional networks for large-scale image recognition.
\newblock \emph{arXiv preprint arXiv:1409.1556} .

\bibitem[{Stamoulis et~al.(2019)Stamoulis, Ding, Wang, Lymberopoulos,
  Priyantha, Liu, and Marculescu}]{stamoulis2019single}
Stamoulis, D.; Ding, R.; Wang, D.; Lymberopoulos, D.; Priyantha, B.; Liu, J.;
  and Marculescu, D. 2019.
\newblock Single-path nas: Designing hardware-efficient convnets in less than 4
  hours.
\newblock \emph{arXiv preprint arXiv:1904.02877} .

\bibitem[{Szegedy et~al.(2015)Szegedy, Liu, Jia, Sermanet, Reed, Anguelov,
  Erhan, Vanhoucke, and Rabinovich}]{szegedy2015going}
Szegedy, C.; Liu, W.; Jia, Y.; Sermanet, P.; Reed, S.; Anguelov, D.; Erhan, D.;
  Vanhoucke, V.; and Rabinovich, A. 2015.
\newblock Going deeper with convolutions.
\newblock In \emph{CVPR}.

\bibitem[{Tan et~al.(2019)Tan, Chen, Pang, Vasudevan, Sandler, Howard, and
  Le}]{tan2019mnasnet}
Tan, M.; Chen, B.; Pang, R.; Vasudevan, V.; Sandler, M.; Howard, A.; and Le,
  Q.~V. 2019.
\newblock Mnasnet: Platform-aware neural architecture search for mobile.
\newblock In \emph{CVPR}.

\bibitem[{Tan and Le(2019)}]{tan2019efficientnet}
Tan, M.; and Le, Q.~V. 2019.
\newblock Efficientnet: Rethinking model scaling for convolutional neural
  networks.
\newblock \emph{arXiv preprint arXiv:1905.11946} .

\bibitem[{Wang et~al.(2020)Wang, Gao, Chen, Wang, Tian, Shen, and
  Zhang}]{wang2020fcos}
Wang, N.; Gao, Y.; Chen, H.; Wang, P.; Tian, Z.; Shen, C.; and Zhang, Y. 2020.
\newblock Nas-fcos: Fast neural architecture search for object detection.
\newblock In \emph{Proceedings of the IEEE/CVF Conference on Computer Vision
  and Pattern Recognition}, 11943--11951.

\bibitem[{Xie et~al.(2019)Xie, Kirillov, Girshick, and He}]{xie2019exploring}
Xie, S.; Kirillov, A.; Girshick, R.; and He, K. 2019.
\newblock Exploring randomly wired neural networks for image recognition.
\newblock \emph{arXiv preprint arXiv:1904.01569} .

\bibitem[{Xie et~al.(2018)Xie, Zheng, Liu, and Lin}]{xie2018snas}
Xie, S.; Zheng, H.; Liu, C.; and Lin, L. 2018.
\newblock SNAS: stochastic neural architecture search.
\newblock \emph{arXiv preprint arXiv:1812.09926} .

\bibitem[{Xu et~al.(2019)Xu, Xie, Zhang, Chen, Qi, Tian, and Xiong}]{xu2019pc}
Xu, Y.; Xie, L.; Zhang, X.; Chen, X.; Qi, G.-J.; Tian, Q.; and Xiong, H. 2019.
\newblock Pc-darts: Partial channel connections for memory-efficient
  differentiable architecture search.
\newblock \emph{arXiv preprint arXiv:1907.05737} .

\bibitem[{Yan et~al.(2020)Yan, Zheng, Ao, Zeng, and Zhang}]{yan2020does}
Yan, S.; Zheng, Y.; Ao, W.; Zeng, X.; and Zhang, M. 2020.
\newblock Does unsupervised architecture representation learning help neural
  architecture search?
\newblock \emph{Advances in Neural Information Processing Systems} 33.

\bibitem[{Yang, Esperan{\c{c}}a, and Carlucci(2019)}]{yang2019evaluation}
Yang, A.; Esperan{\c{c}}a, P.~M.; and Carlucci, F.~M. 2019.
\newblock NAS evaluation is frustratingly hard.
\newblock \emph{arXiv preprint arXiv:1912.12522} .

\bibitem[{Yang et~al.(2020)Yang, Li, You, Wang, Qian, and Lin}]{yang2020ista}
Yang, Y.; Li, H.; You, S.; Wang, F.; Qian, C.; and Lin, Z. 2020.
\newblock Ista-nas: Efficient and consistent neural architecture search by
  sparse coding.
\newblock \emph{Advances in Neural Information Processing Systems} 33.

\bibitem[{You et~al.(2020)You, Huang, Yang, Wang, Qian, and
  Zhang}]{you2020greedynas}
You, S.; Huang, T.; Yang, M.; Wang, F.; Qian, C.; and Zhang, C. 2020.
\newblock GreedyNAS: Towards Fast One-Shot NAS with Greedy Supernet.
\newblock In \emph{Proceedings of the IEEE/CVF Conference on Computer Vision
  and Pattern Recognition}, 1999--2008.

\bibitem[{Zela et~al.(2020)Zela, Elsken, Saikia, Marrakchi, Brox, and
  Hutter}]{zela2019understanding}
Zela, A.; Elsken, T.; Saikia, T.; Marrakchi, Y.; Brox, T.; and Hutter, F. 2020.
\newblock Understanding and Robustifying Differentiable Architecture Search.
\newblock \emph{ICML} .

\bibitem[{Zela, Siems, and Hutter(2020)}]{zela2020bench}
Zela, A.; Siems, J.; and Hutter, F. 2020.
\newblock NAS-Bench-1Shot1: Benchmarking and Dissecting One-shot Neural
  Architecture Search.
\newblock \emph{arXiv preprint arXiv:2001.10422} .

\bibitem[{Zhong et~al.(2020)Zhong, Deng, Guo, Scott, and
  Huang}]{zhong2020representation}
Zhong, Y.; Deng, Z.; Guo, S.; Scott, M.~R.; and Huang, W. 2020.
\newblock Representation Sharing for Fast Object Detector Search and Beyond.
\newblock \emph{arXiv preprint arXiv:2007.12075} .

\bibitem[{Zoph and Le(2016)}]{zoph2016neural}
Zoph, B.; and Le, Q.~V. 2016.
\newblock Neural architecture search with reinforcement learning.
\newblock \emph{arXiv preprint arXiv:1611.01578} .

\bibitem[{Zoph et~al.(2018)Zoph, Vasudevan, Shlens, and Le}]{zoph2018learning}
Zoph, B.; Vasudevan, V.; Shlens, J.; and Le, Q.~V. 2018.
\newblock Learning transferable architectures for scalable image recognition.
\newblock In \emph{CVPR}.

\end{thebibliography}

\clearpage
\section{Appendix}
In this appendix, we provide additional material to supplement our main submission, including additional ablation study, complexity analysis, datasets, implementation details and visualization of cell-level structures and stage-level structures in $H^c$-DAS and $H^s$-DAS.

\subsection{Ablation Study}
\begin{figure*}[htb]
  \begin{center}
    \subfigbottomskip=30pt
  \subfigure[w/o Sliding Window, w/o Depth Loss]{
    \label{depth-1}
    \includegraphics[width=0.43\textwidth]{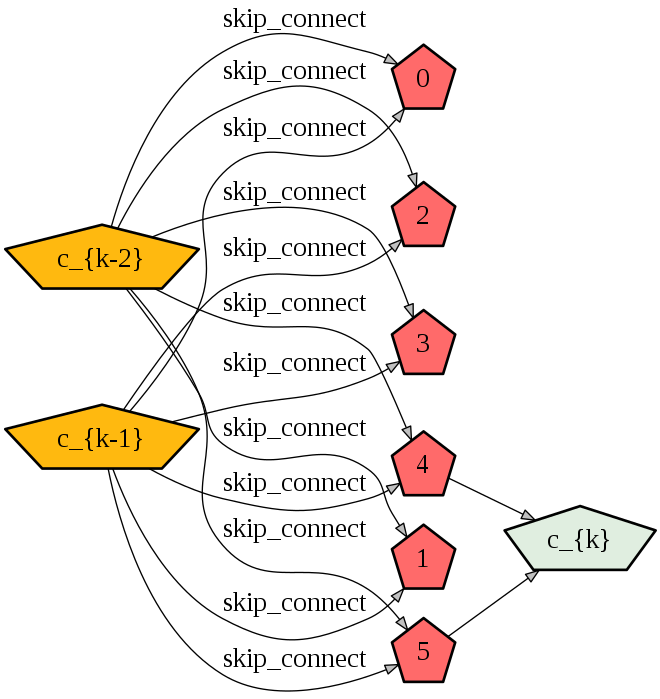}
  }
  \quad
  \subfigure[Sliding Window, w/o Depth Loss]{
    \label{depth-2}
    \includegraphics[width=0.68\textwidth]{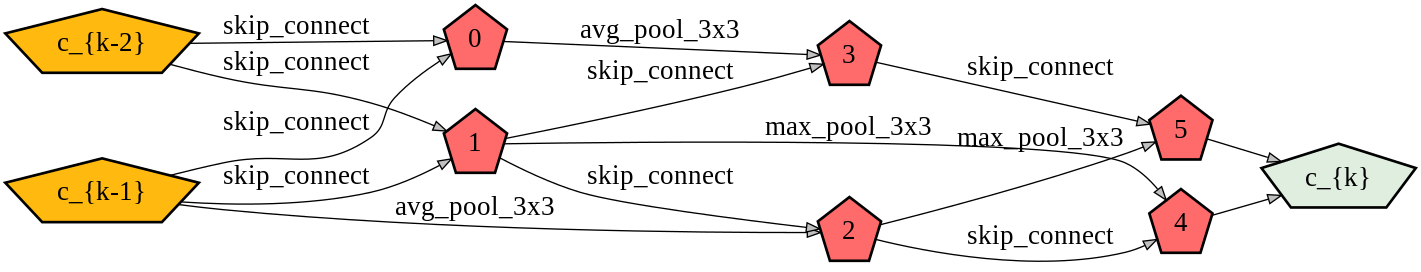}
  }
  \quad
  \subfigure[Sliding Window, Depth Loss]{
    \label{depth-3}
    \includegraphics[width=0.98\textwidth]{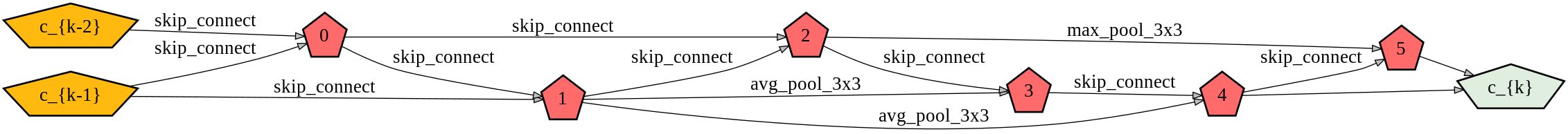}
  }
  \end{center}
  \caption{The depths of the stage-level structures are increasing through adding the sliding window (the input restrictions of cells) and the depth loss.}
  \label{fig:depth}
\end{figure*}

\paragraph{Depth loss.} Once the number of cells in each stage is determined, we then fix the number of cells, and focus on the search of stage architectures. The loss function of the search period is $Loss = L_{cls} + \delta L_{depth}$.
In order to show the importance of the depth loss, we do some ablation experiments on CIFAR10. As we see in Table~\ref{tab:delta}, the depth loss brings a great improvement on the performance. We show the stage-level structure searched without sliding window and depth loss in fig.\ref{depth-1}, structure searched with sliding window but without depth loss in fig.\ref{depth-2} and that searched with both sliding window and depth loss in fig.\ref{depth-3}. We define the depth of an architecture as the average length of the paths from input to output, and obviously, the depths of the three stage-level structures are increasing.

\noindent\textbf{Connections and operations. }
We have a key observation that the connections between cells may play a more important role than the operation types. We show an example in Table~\ref{tab:diff}. Therefore, in stage-level structures, we only choose several non-parametric operations.

\subsection{Complexity Analysis}
In this section, we analysis the complexity of our search space for $H^c$-DAS and $H^s$-DAS.

In DARTS, each of the discretized cell allows $\prod_{k=1}^4\frac{(k+1)k}{2}\times(7^2)\approx10^9$ possible DAGs without considering graph isomorphism.
Since they are jointly learning both normal and reduction cells, the total number of architectures is approximately $(10^9)^2=10^{18}$.

In $H^c$-DAS, we have three stage-specific normal cells and two different reduction cells, so the total number of architectures is approximately $(10^9)^5=10^{45}$.

In $H^s$-DAS, each of our stage-level structure allows $\prod_{k=1}^6\frac{(k+1)k}{2}\times(3^2)\approx10^{10}$ possible DAGs without considering graph isomorphism (recall we have 3 non-zero ops, 2 input nodes, 6 intermediate cells with 2 predecessors each).  If we introduce the sliding window constraint to the search space, it becomes $(\prod_{k=1}^2\frac{(k+1)k}{2}\times(3^2)\prod_{k=3}^6\frac{2\times 3}{2}\times(3^2))^3\approx10^{24}$.
Combined with the cell-level structures, the total number of architectures for CIFAR10 is $(10^{10})^3\times(10^{9})^2=10^{42}$.

The search space is even larger in the cell allocation search, as there are 8 cells per stage. Hence, each of the stage-level structure has a search space of $\prod_{k=1}^4\frac{(k+1)k}{2}\times(3^2)+\prod_{k=1}^5\frac{(k+1)k}{2}\times(3^2)+\prod_{k=1}^6\frac{(k+1)k}{2}\times(3^2)+\prod_{k=1}^7\frac{(k+1)k}{2}\times(3^2)+\prod_{k=1}^8\frac{(k+1)k}{2}\times(3^2)\approx10^{15}$. This results in a total number of $(10^{15})^3=10^{45}$ for the whole network.
If we introduce the sliding window constraint to the search space, the search space becomes $(\prod_{k=1}^2\frac{(k+1)k}{2}\times(3^2)\prod_{k=3}^4\frac{2\times 3}{2}\times(3^2)+\prod_{k=1}^2\frac{(k+1)k}{2}\times(3^2)\prod_{k=3}^5\frac{2\times 3}{2}\times(3^2)+\prod_{k=1}^2\frac{(k+1)k}{2}\times(3^2)\prod_{k=3}^6\frac{2\times 3}{2}\times(3^2)+\prod_{k=1}^2\frac{(k+1)k}{2}\times(3^2)\prod_{k=3}^7\frac{2\times 3}{2}\times(3^2)+\prod_{k=1}^2\frac{(k+1)k}{2}\times(3^2)\prod_{k=3}^8\frac{2\times 3}{2}\times(3^2))^3\approx10^{33}$.
Combined with the cell-level structures, the total number of architectures for CIFAR10 is about $10^{51}$.

\begin{table}[t]
  \centering
  \begin{tabular}{c|c}
  \hline
  $\delta$ & Test Err. on CIFAR10 (\%)\\ \shline
  0     & 2.59          \\
  0.33  & 2.47          \\
  1     & 2.34          \\
  1.5   & 2.53          \\\hline
  \end{tabular}
  \caption{Depth loss. $\delta$ is used to balance the classification loss and depth loss.}
  \label{tab:delta}
\end{table}

\subsection{Datasets}
We perform experiments on CIFAR10~\cite{krizhevsky2009learning} and ImageNet~\cite{deng2009imagenet}, which are two image classification benchmarks for evaluating neural architecture search.

CIFAR10 consists of 50K training images and 10K testing images. These images are of a spatial resolution of $32\times32$ and equally distributed over 10 classes.
ImageNet is a large-scale and well known benchmark for image classification. It contains 1000 object categories, 1.28M training images, and 50K validation images. Following the ~\cite{liu2018darts}, we apply the setting that the input size is fixed to be $224\times224$. 

\begin{table}[t]
  \centering
  \begin{tabular}{ccc}
  \hline
  operations                        & stage-level structure                  & top-1              \\ \shline
  \multicolumn{1}{c|}{\textit{none}}         & \multicolumn{1}{c|}{\multirow{4}{*}{\includegraphics[width=0.23\textwidth]{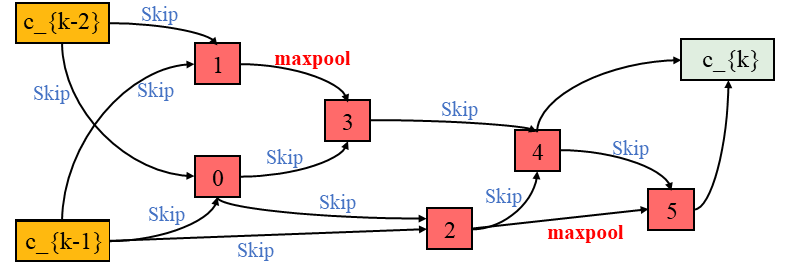}}} & \multirow{4}{*}{97.49} \\
  \multicolumn{1}{c|}{\textit{skip-connect}} & \multicolumn{1}{c|}{}                  &                        \\
  \multicolumn{1}{c|}{\textit{max-pool-3x3}} & \multicolumn{1}{c|}{}                  &                        \\
  \multicolumn{1}{c|}{\textit{avg-pool-3x3}} & \multicolumn{1}{c|}{}                  &                        \\ \hline
  \multicolumn{1}{c|}{}             & \multicolumn{1}{c|}{\multirow{4}{*}{\includegraphics[width=0.23\textwidth]{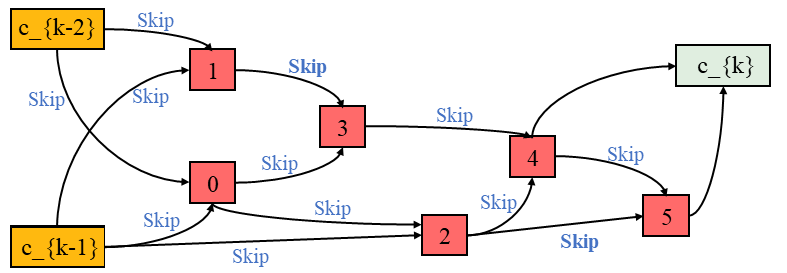}}} & \multirow{4}{*}{97.48} \\
  \multicolumn{1}{c|}{\textit{none}}         & \multicolumn{1}{c|}{}                  &                        \\
  \multicolumn{1}{c|}{\textit{skip-connect}} & \multicolumn{1}{c|}{}                  &                        \\
  \multicolumn{1}{c|}{}             & \multicolumn{1}{c|}{}                  &                        \\ \hline
  \end{tabular}
  \caption{Connection and Operations. In stage-level search space, two structures share the same connection, only with some operations changed. We evaluate each of the structures three times on CIFAR10 and show the average accuracy, which are almost equal.}
  \label{tab:diff}
\end{table}

\subsection{Implementation Details}
\paragraph{Architecture search in cell-level search space.}
The entire experiments have three stages: architecture search for cells, architecture search for macro-architectures and architecture evaluation.
Both $H^{c}$-DAS and $H^{s}$-DAS need architecture search for cells and architecture evaluation, and $H^{s}$-DAS needs to search for macro-architecture additionally.
The search space for cells is the same as DARTS, which has 8 candidate operations including:

- $3\times3$ depthwise-separable conv    ~ - $3\times3$ average pooling

- $5\times5$ depthwise-separable conv     - $3\times3$ max pooling

- $3\times3$ dilated conv  ~~~~~~~~~~~~~~~~~~~~~ - skip connection

- $5\times5$ dilated conv ~~~~~~~~~~~~~~~~~~~~~~- no connection (none) 

Meanwhile, the search space for macro-architecture has only 4 non-parametric candidate operations including:

- $3\times3$ average pooling~~~~~~~~~~~~~~~~~-skip connection

- $3\times3$ max pooling  ~~~~~~~~~~~~~~~~~~~~~- no connection(none)

We have narrowed down the choice of stage-level operations because we don't want to bring too much learnable architecture parameters to $H^{s}$-DAS. 

For CIFAR10, in $H^{c}$-DAS we use the same structure as these one-shot methods by stacking 8 cells (6 normal cells and 2 reduction cells). 
In our method, the 8 layers network is separated into three stages by 2 different reduction cells. Each stage contains 2 normal cells, and the normal cells in different stages are different. The whole network is stacked by 2 \textit{normal cells} in the first stage, 1 \textit{reduction cell}, 2 \textit{normal cells} in the second stage, 1 \textit{reduction cell} and 2 \textit{normal cells} in the third stage. Each cell consists of $N=6$ nodes. We search for the architecture of cells for 50 epochs with initial 16 channels and 64 batch size. Half of the training data is used to update the model weights while the other is used to update the architecture parameters. The model weights are optimized by momentum SGD with initial learning rate of 0.025, momentum of 0.9 and weight decay of $2.7\times10^{-3}$. The architecture parameters are optimized by Adam with a fixed learning rate of $3\times10^{-4}$, a momentum (0.5, 0.999) and a weight decay of $10^{-3}$.

For ImageNet, the one-shot model starts with three $3\times3$ convolution layers with stride 2 to reduce the resolution of input images from $224\times224$ to $28\times28$.
The cells we use are derived on CIFAR10, which are shown in figure~\ref{fig:nc} and \ref{fig:rc}.

\noindent\textbf{Architecture search in stage-level search space. }
For the macro-architectures searching on CIFAR10, the initial channel is 16, batch size is 64 and the number of training epochs is 50.
The optimizers of model weights and architecture parameters are SGD and Adam respectively. The setting of optimizers is just the same as that in the period of searching for cells. In order to construct a network of 20 cells, we search for three macro-structures each of which has 6 normal cells.

For the macro-architectures searching on ImageNet, the cells are derived on CIFAR10.  The initial channel is 16, the batch size is 64 and the training epochs is 50. Considering the evaluation stage on ImageNet with 14 stacked cells, we should search three macro-architectures that each stage contains four normal cells. Each cell in a macro-architecture can only choose its preceding cells from three previous cells. Due to the difficulty of bi-level optimization on ImageNet, we search the macro-architectures on CIFAR10 and split half of it for updating model weights and another half for updating architecture parameters of macro-architectures. The model weights are optimized by SGD with initial learning rate of 0.025, momentum of 0.9 and weight decay of $2.7\times10^{-3}$. The architecture parameters of macro-architectures are optimized by Adam with a fixed learning rate of $3\times10^{-4}$, a momentum (0.5, 0.999) and a weight decay of $10^{-3}$. 
We visualize the macro-architectures of CIFAR10 in fig.~\ref{fig:3dag-cifar} and those of ImageNet in fig.~\ref{fig:3dag-img}.

In short, for a fair comparison, we inherit the configuration of the original DARTS as far as possible. For CIFAR10, the searching time of cells is 0.4 GPU days with a single Titan X and the searching time of macro-architectures is 0.3 GPU days. For ImageNet, we transfer the cells searched on CIFAR10, and just search for macro-architectures in 0.3 GPU days. 

\noindent\textbf{Architecture evaluation. }
The evaluation stage of $H^c$-DAS is the same as that of DATRS.
Considering $H^s$-DAS for CIFAR10, the network is composed of 20 cells, and six same normal cells construct a macro-architecture. The initial number of channels is 36 and the total training images are used. The network is trained from scratch for 600 epochs with a batch size of 128. We use the SGD optimizer with an initial learning rate of 0.025, a momentum of 0.9, a weight decay of $3\times10^{-4}$ and a gradient clipping for weights of 5. Auxiliary weight of 0.4 and drop path probability of 0.2 are added for regularization. 

The evaluation stage of ImageNet also starts with three convolution layers of stride 2 to reduce the resolution from $224\times224$ to $28\times28$. The network stacks three macro-architectures with two reduction cells, and each stage contains four normal cells. In order to keep the number of multiply-add operations under mobile setting, the initial channel number is changed to be 39 and the training epoch number is 250 with a batch size of 128. We use the SGD optimizer with a momentum of 0.9, an initial learning rate of 0.1 and a weight decay of $3\times10^{-5}$. Additional enhancements are adopted including label smoothing and an auxiliary loss tower with a weight of 0.4.

\subsection{Visualization}

\noindent\textbf{Colors.}
When plotting cell-level structures, we use the colors of darkseagreen2, lightblue and palegoldenrod, while plotting stage-level structures, the colors are darkgoldenrod1, indianRed1 and honeydew2, which represent input features, nodes (or cells in stage-level structures) and output feature respectively.

\begin{figure*}[htb]
  \begin{center}
    \subfigbottomskip=30pt
  \subfigure[normal cell in the 1st stage]{
    \label{3d3c.nc.1}
    \includegraphics[width=0.45\textwidth]{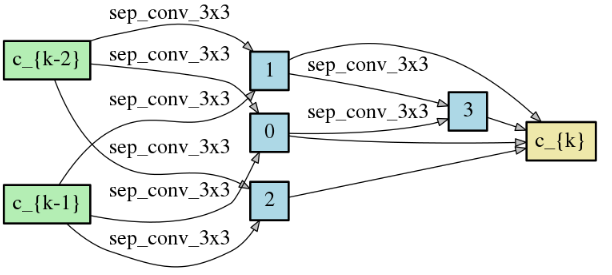}
  }
  \quad
  \subfigure[normal cell in the 2nd stage]{
    \label{3d3c.nc.2}
    \includegraphics[width=0.32\textwidth]{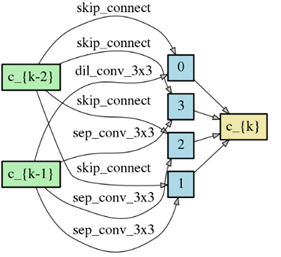}
  }
  \quad
  \subfigure[normal cell in the 3rd stage]{
    \label{3d3c.nc.3}
    \includegraphics[width=0.65\textwidth]{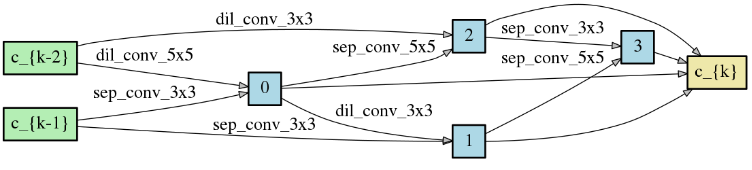}
  }
  \end{center}
  \caption{The best normal cells of $H^{c}$-DAS searched on CIFAR10.}
  \label{fig:nc}
\end{figure*}

\begin{figure*}[htb]
  \begin{center}

  \subfigure[reduction cell 1]{
  \label{3d3c.rc.1}
  \includegraphics[width=0.53\textwidth]{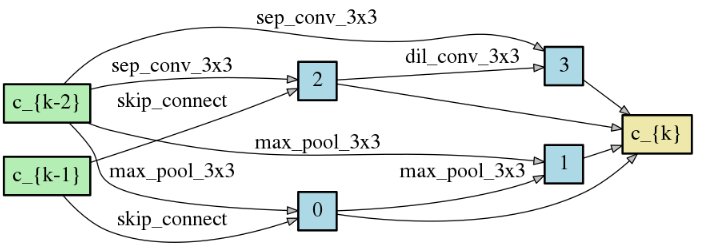}
  }
  \hfill
  \subfigure[reduction cell 2]{
  \label{3d3c.rc.2}
  \includegraphics[width=0.83\textwidth]{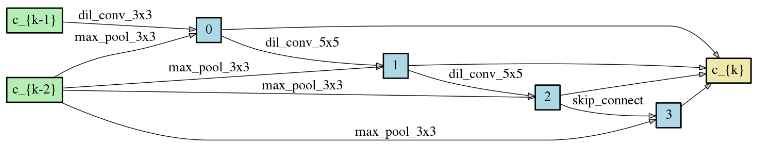}
  }
  
  \end{center}
    \caption{The best reduction cells of $H^{c}$-DAS searched on CIFAR10.}
  \label{fig:rc}
\end{figure*}

\noindent\textbf{Cell-level structures.}
In Fig~\ref{fig:nc}, we visualize the three stage-specific normal cells of $H^c$-DAS, and Fig~\ref{fig:rc} shows the reduction cells. All of them are searched on CIFAR10 and derived on CIFAR10 and ImageNet in $H^c$-DAS. An interesting discovery is that the normal cells in the 1st stage (fig.\ref{3d3c.nc.1}) are full of convolution with small kernels ($3\times3$), while the normal cells in the 3rd stage (fig.\ref{3d3c.nc.3}) prefer convolution with various kernels ($3\times3$ and $5\times5$).

\begin{figure*}[htb]
  \begin{center}
  \subfigure[the 1st macro-architecture]{
  \label{3d3c.bd.1}
  \includegraphics[width=0.85\textwidth]{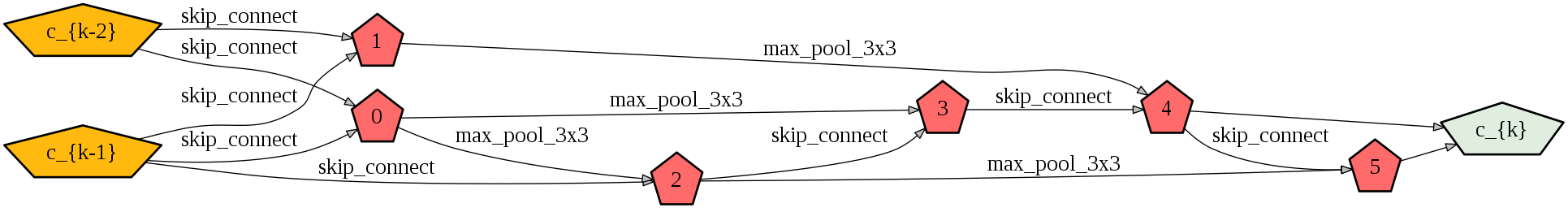}
  }
  \hfill
  \subfigure[the 2nd macro-architecture]{
  \label{3d3c.bd.2}
  \includegraphics[width=0.85\textwidth]{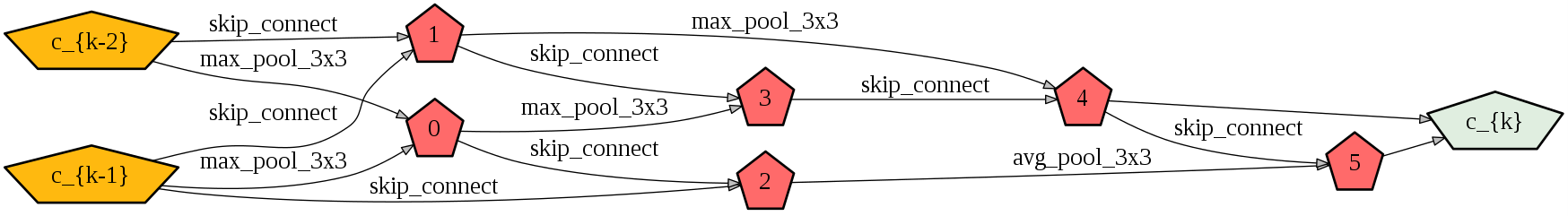}
  }
  \hfill
  \subfigure[the 3rd macro-architecture]{
  \label{3d3c.bd.3}
  \includegraphics[width=0.85\textwidth]{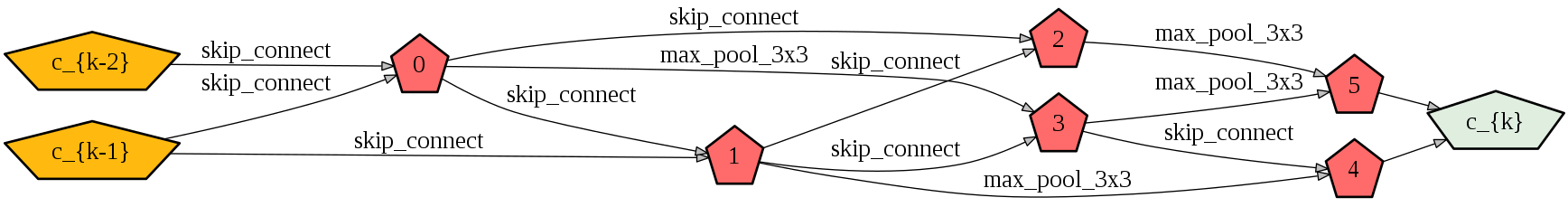}
  }
  
  \end{center}
    \caption{Stage-level Structures of $H^s$-DAS for CIFAR10. Each stage has 6 cells.}
  \label{fig:3dag-cifar}
  \end{figure*}

\begin{figure*}[htb]
    \begin{center}
    \subfigure[the 1st macro-architecture]{
    \includegraphics[width=0.85\textwidth]{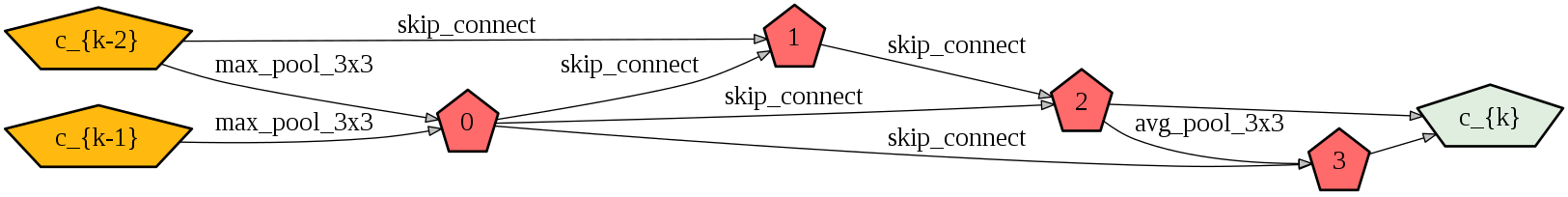}
    }
    \hfill
    \subfigure[the 2nd macro-architecture]{
    \includegraphics[width=0.85\textwidth]{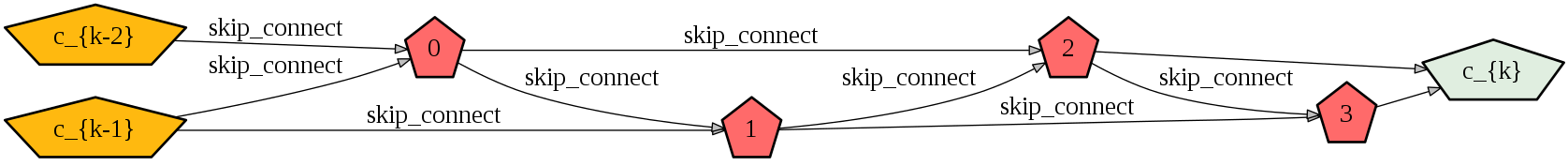}
    }
    \hfill
    \subfigure[the 3rd macro-architecture]{
    \includegraphics[width=0.85\textwidth]{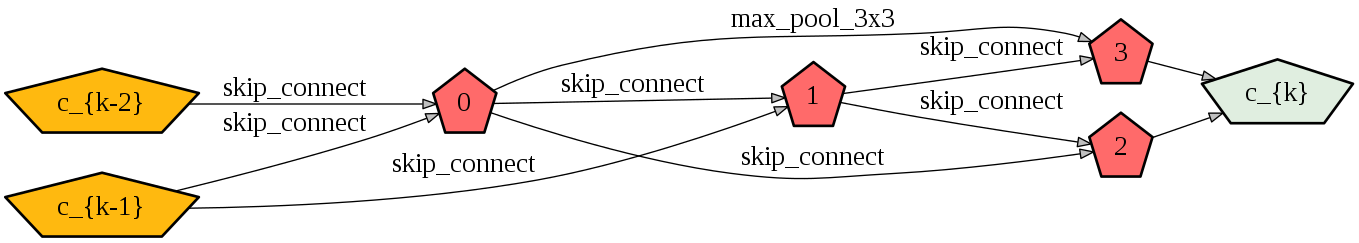}
    }
    
    \end{center}
      \caption{Stage-level Structures of $H^s$-DAS for ImageNet. Each stage has 4 cells.}
    \label{fig:3dag-img}
\end{figure*}

\noindent\textbf{Stage-level structures.}
There are non-parametric connections between cells in stage-level structures of $H^s$-DAS. In Fig.~\ref{fig:3dag-cifar}, we construct a 20-cell network for CIFAR10, each stage of which has 6 cells. Because of the input restrictions of cells, each cell could connect with 3 candidate preceding cells in its stage.
In Fig.~\ref{fig:3dag-img}, we construct a 14-cell network for ImageNet, each stage of which has 4 cells.

\begin{figure*}[htb]
  \begin{center}
  \subfigure[the 1st macro-structure for CIFAR10]{
  \includegraphics[width=0.95\textwidth]{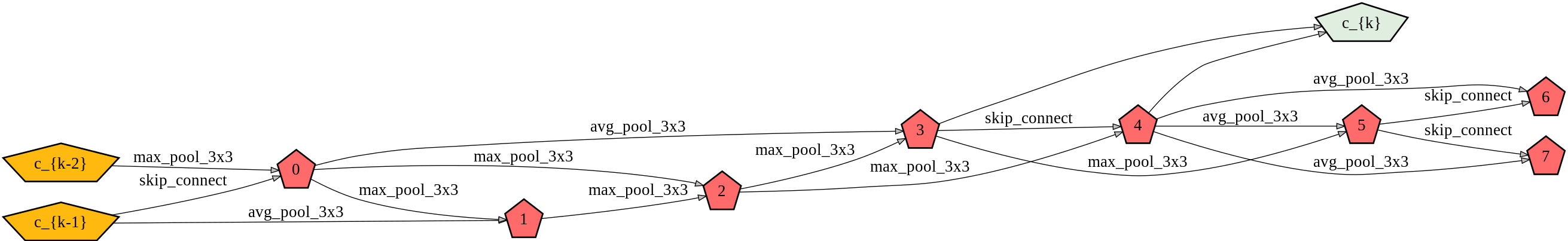}
  }
  \hfill
  \subfigure[the 2nd macro-structure for CIFAR10]{
  \includegraphics[width=0.95\textwidth]{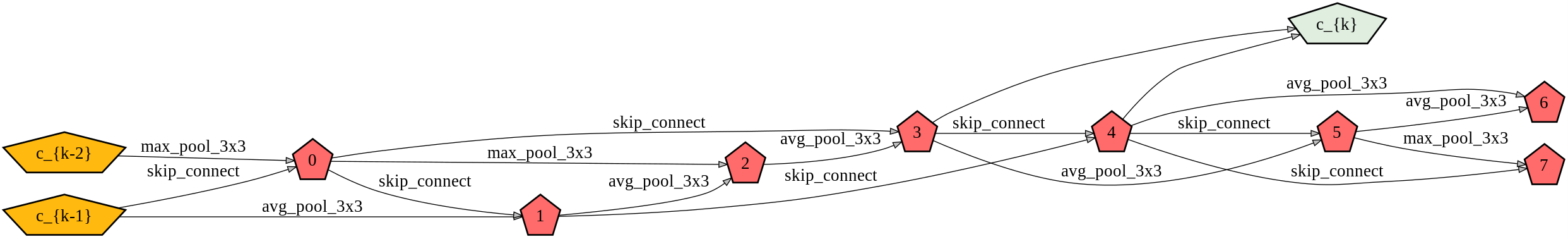}
  }
  \hfill
  \subfigure[the 3rd macro-structure for CIFAR10]{
  \includegraphics[width=0.95\textwidth]{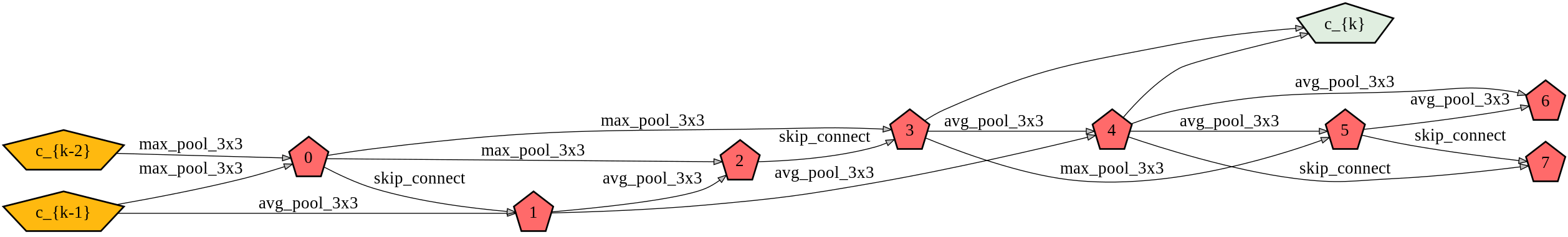}
  }
  
  \end{center}
    \caption{Larger $\gamma$ introduces less cells to stage-level structures in $H^s$-DAS. The cell distribution remains the same.}
  \label{fig:bigPL}
\end{figure*}

\begin{figure*}[htb]
  \begin{center}
  \subfigure[the 1st macro-structure for CIFAR10]{
  \includegraphics[width=0.95\textwidth]{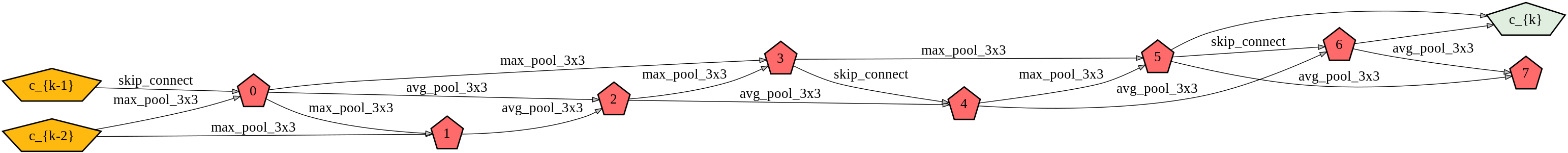}
  }
  \hfill
  \subfigure[the 2nd macro-structure for CIFAR10]{
  \includegraphics[width=0.95\textwidth]{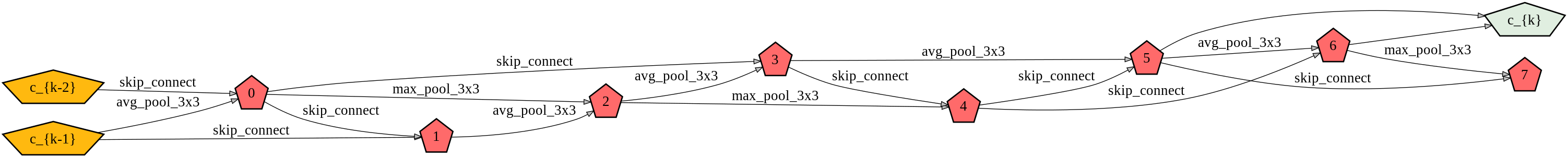}
  }
  \hfill
  \subfigure[the 3rd macro-structure for CIFAR10]{
  \includegraphics[width=0.95\textwidth]{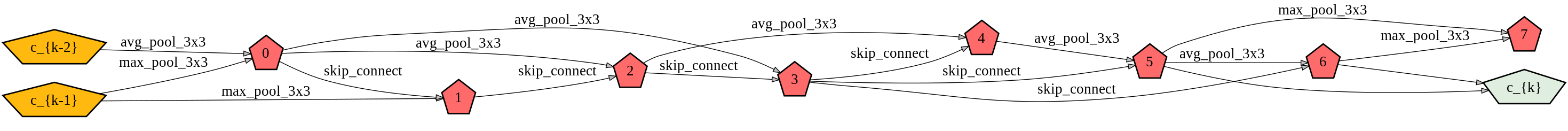}
  }

  \end{center}
    \caption{Smaller $\gamma$ introduces more cells to stage-level structures in $H^s$-DAS. The cell distribution remains the same.}
  \label{fig:smallPL}
\end{figure*}

\noindent\textbf{Loss of computational complexity.}
The search for the distribution of cells over three stages is conducted under a constraint of certain computational complexity. Adjusting the weighting factor $\gamma$ in Equation $Loss = L_{cls} + \delta L_{depth} + \gamma\times L_{comp}$ leads to different total number of cells in the network, but the cell distribution remains the same, which means the number of cells in each stage is the same at the end of the search. In Fig.~\ref{fig:bigPL} and Fig.~\ref{fig:smallPL}, we show how the weighting factor $\gamma$ changes the number of cells in each stage.

\end{document}